%% file: root.tex
\documentclass[letterpaper, 10 pt, conference]{ieeeconf}  % Comment this line out if you need a4paper

\IEEEoverridecommandlockouts                              % This command is only needed if 
\usepackage{amsmath} % assumes amsmath package installed
\usepackage{amssymb}  % assumes amsmath package installed
\usepackage{graphicx}
\usepackage{adjustbox}
\usepackage[font=footnotesize, format=plain, labelfont=bf]{caption}
\usepackage{hyperref}

\usepackage[printonlyused]{acronym}
\input{chapters/00_acronyms}
\usepackage[dvipsnames]{xcolor}
\usepackage{fontawesome}
\usepackage{array}
\usepackage{tabularx}
\newcolumntype{K}[1]{>{\RaggedRight\arraybackslash}p{#1}}
\usepackage{booktabs}
\usepackage{multirow}
\usepackage{siunitx}
\usepackage{ragged2e}
\usepackage{pifont}

\usepackage[nocompress]{cite}


\usepackage{tikz}
\newcommand*\emptycirc{\tikz\draw (0,0) circle (1.0ex);} 
\newcommand*\halfcirc{%
  \tikz[baseline=(moon.base)]{
    \draw (0,0) circle (1ex); \path[fill] (0,0) -- (90:1ex) arc (90:270:1ex) -- cycle; \coordinate (moon) at (0,-1ex);
  }
}
\newcommand*\fullcirc{\tikz\fill (0,0) circle (1.0ex);} 

\begin{document}

\title{\LARGE \bf
A Practical Framework of Key Performance Indicators for \\ Multi-Robot Lunar and Planetary Field Tests
}

\author{Julia Richter$^{*,1}$, David Oberacker$^{*,2,3}$, Gabriela Ligeza$^{4,5}$, Valentin T. Bickel$^6$, Philip Arm$^1$, \\ 
William Talbot$^1$, Marvin Grosse Besselmann$^2$, Florian Kehl$^{7,8}$, Tristan Schnell$^2$, \\
Hendrik Kolvenbach$^1$, Rüdiger Dillmann$^2$, Arne Roennau$^{2,3}$, and Marco Hutter$^1$%
\thanks{$^*$ Equal contribution}%
\thanks{$^1$ Robotic Systems Lab (RSL), ETH Zürich, Zürich, Switzerland;
$^2$ FZI Research Center for Information Technology, Karlsruhe, Germany;
$^3$ Machine Intelligence and Robotics Lab (MaiRo), Karlsruhe Institute for Technology (KIT), Karlsruhe, Germany;
$^4$ Department of Environmental Sciences, University of Basel, Basel, Switzerland;
$^5$ European Space Agency/ESTEC, Noordwijk, Netherlands;
$^6$ Center for Space and Habitability, University of Bern, Switzerland;
$^7$ Space Instruments Group, University of Zürich, Zürich, Switzerland;
$^8$ Space Science and Technology, ETH Zürich, Zürich, Switzerland}
}

\maketitle
\pagestyle{empty}

%%%%%%%%%%%%%%%%%%%%%%%%%%%%%%%%%%%%%%%%%%%%%%%%%%%%%%%%%%%%%%%%%%%%%%%%%%%%%%%%
\begin{abstract}
\input{chapters/00_abstract}


\end{abstract}

%%%%%%%%%%%%%%%%%%%%%%%%%%%%%%%%%%%%%%%%%%%%%%%%%%%%%%%%%%%%%%%%%%%%%%%%%%%%%%%%
\section{INTRODUCTION}
\label{sec:INTRODUCTION}
\input{chapters/03_mission_scenarios_fig1}
\input{chapters/01_introduction}

\section{RELATED WORK}
\label{sec:RELATED WORK}
\input{chapters/02_related_work}

\section{MISSION SCENARIOS}
\label{sec:MISSION SCENARIOS}
\input{chapters/03_mission_scenarios}

\section{KPI DEFINITION}
\label{sec:KPI DEFINITION}
\input{chapters/04_kpi_definition}

\section{DISCUSSION}
\label{sec:DISCUSSION}
\input{chapters/05_discussion}

\section{CONCLUSION}
\label{sec:CONCLUSION}
\input{chapters/06_conclusion}

%%%%%%%%%%%%%%%%%%%%%%%%%%%%%%%%%%%%%%%%%%%%%%%%%%%%%%%%%%%%%%%%%%%%%%%%%%%%%%%%

%%%%%%%%%%%%%%%%%%%%%%%%%%%%%%%%%%%%%%%%%%%%%%%%%%%%%%%%%%%%%%%%%%%%%%%%%%%%%%%%

%%%%%%%%%%%%%%%%%%%%%%%%%%%%%%%%%%%%%%%%%%%%%%%%%%%%%%%%%%%%%%%%%%%%%%%%%%%%%%%%

\section*{ACKNOWLEDGMENT}
This work was supported by the European Space Agency (ESA) (Ref. 4000141520/23/NL/AT), the DLR Space Administration under grant agreement No. 50RA2404 by the German Bundestag, the Luxembourg National Research Fund (Ref. 18990533), and the Swiss National Science Foundation (SNSF) as part of the projects No.200021E\_229503 and No.227617.

%%%%%%%%%%%%%%%%%%%%%%%%%%%%%%%%%%%%%%%%%%%%%%%%%%%%%%%%%%%%%%%%%%%%%%%%%%%%%%%%

\bibliographystyle{IEEEtran}
\bibliography{bibtex}

\end{document}

%% file: chapters/00_acronyms.tex
\acrodef{ISRU}{In Situ Resource Utilization}
\acrodef{ESA}{European Space Agency}
\acrodef{NASA}{National Aeronautics and Space Administration}
\acrodef{KPI}{Key Performance Indicator}
\acrodef{HRI}{Human-Robot Interaction}
\acrodef{REE}{Rare Earth Element}
\acrodef{PSR}{Permanently Shadowed Region}
\acrodef{LRO}{Lunar Reconnaissance Orbiter}
\acrodef{XRF}{X-Ray Fluorescence Spectrometer}
\acrodef{LIBS}{Lased-Induced Breakdown Spectrometer}
\acrodef{APXS}{Alpha Particle X-ray Spectrometer}
\acrodef{GPR}{Ground Penetrating Radar}

%% file: chapters/00_abstract.tex
Robotic prospecting for critical resources on the Moon, such as ilmenite, rare earth elements, and water ice, requires robust exploration methods given the diverse terrain and harsh environmental conditions.
Although numerous analog field trials address these goals, comparing their results remains challenging because of differences in robot platforms and experimental setups.
These missions typically assess performance using selected, scenario-specific engineering metrics that fail to establish a clear link between field performance and science-driven objectives.
In this paper, we address this gap by deriving a structured framework of \acp{KPI} from three realistic multi-robot lunar scenarios reflecting scientific objectives and operational constraints.
Our framework emphasizes scenario-dependent priorities in efficiency, robustness, and precision, and is explicitly designed for practical applicability in field deployments.
We validated the framework in a multi-robot field test and found it practical and easy to apply for efficiency- and robustness-related \acp{KPI}, whereas precision-oriented \acp{KPI} require reliable ground-truth data that is not always feasible to obtain in outdoor analog environments.
Overall, we propose this framework as a common evaluation standard enabling consistent, goal-oriented comparison of multi-robot field trials and supporting systematic development of robotic systems for future planetary exploration.

%% file: chapters/03_mission_scenarios_fig1.tex
\begin{figure}
    \centering
    \includegraphics[width=.9\linewidth]{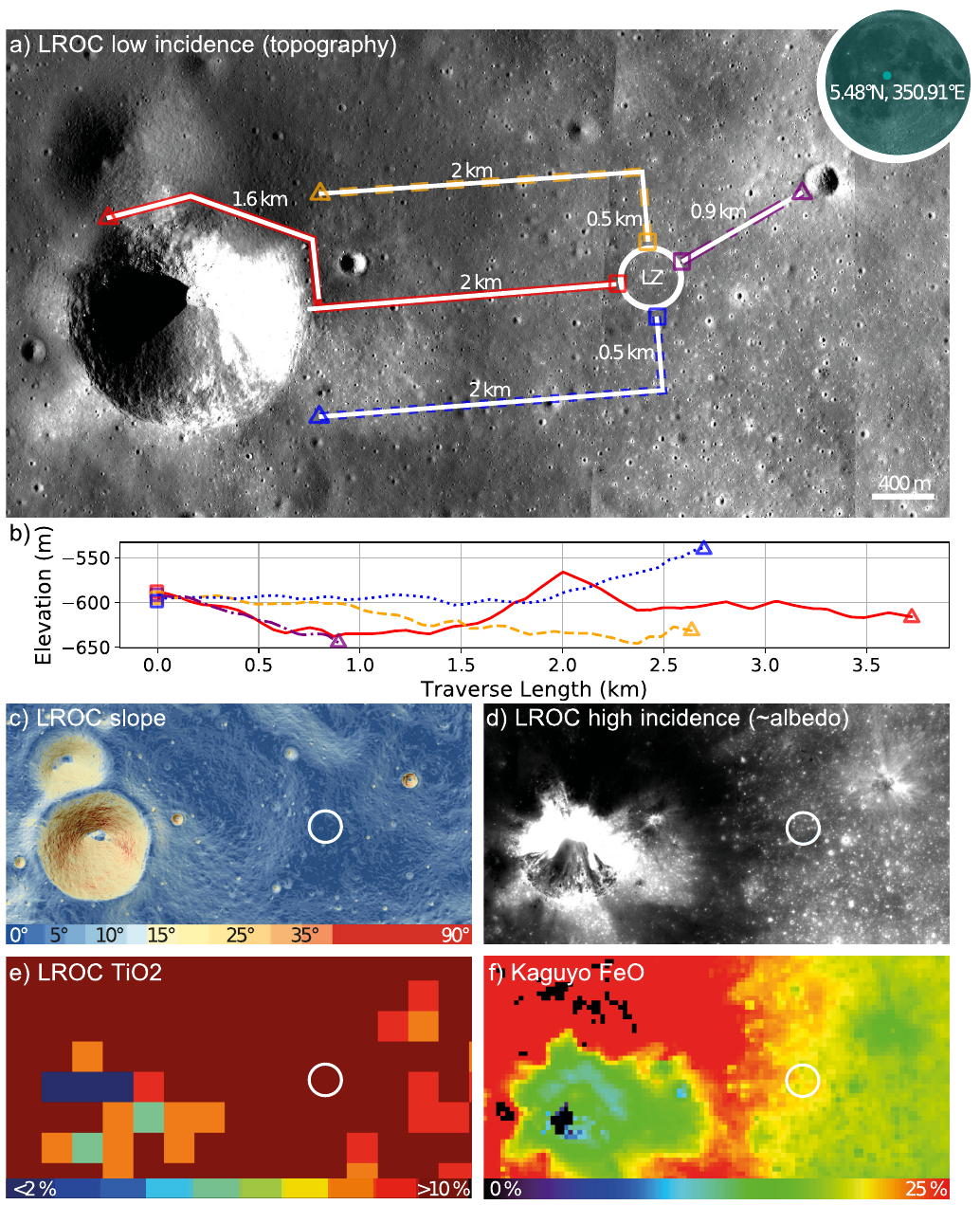}
    \caption{Scenario 1 (Ilmenite). (a) LROC (Lunar Reconnaissance Orbiter) low-incidence image mosaic with landing zone (LZ), preliminary prospecting grid (white), and elevation transect (orange, red, blue, purple). (b) Elevation profile along the transect. (c–f) Selected orbital maps showcasing thermophysical, topographic, and compositional properties, with the LZ indicated by a white circle.}
    \label{fig:scenario1}
\end{figure}

%% file: chapters/01_introduction.tex
\noindent As interest grows in supporting long-term, sustainable presence on the Moon, robotic resource prospecting is becoming an increasingly important research area.
This pursuit is underscored by \acs{ESA}'s Space Resource Strategy \cite{ESA_SpaceResourcesStrategy_2019}, which aims to enable sustainable exploration by locating and characterizing lunar regions expected to contain resources, such as permanently shadowed regions and pyroclastic deposits.
However, reaching and characterizing these locations and resources remains technically challenging. The lunar surface features a wide range of terrain types, from smooth mare plains to steep crater walls and pits, each of which poses different requirements on robotic mobility, sensing capabilities, and autonomy.

Traditional approaches to planetary surface exploration typically rely on single robotic platforms, mostly wheeled rovers, such as Lunokhod 2, Yutu 2, Curiosity, or Perseverance.
Although wheeled platforms perform well on flat terrain with minor obstacles, their performance is limited in more extreme terrain, such as unconsolidated regolith \cite{david2005opportunity}, steep slopes \cite{potts2015robotic}, and other unstructured environments.
Recently, legged platforms have emerged as a viable alternative for traversing such terrains, as demonstrated
for space analog environments \cite{arm2023scientific, kolvenbach2022traversing}. However, this improved mobility comes at the cost of greater energy consumption, mechanical complexity, and control complexity, all of which increase the risk of failure.

Heterogeneous robotic teams that combine different locomotion and sensing capabilities can leverage both approaches to minimize this risk.
Interest in such teams has been fueled by recent competitions, including the DARPA SubT \cite{DoD_Subterranean_Challenge} and the ESA–ESRIC Space Resources Challenge \cite{esa_src_challenge2021}.
Despite imposing no restrictions on the architecture of the system, several teams in the Space Resources Challenge chose a heterogeneous, multi‑robot solution \cite{arm2023scientific, schnell2023efficient, SpaceApplications2022LUVMI}. 
This offered three main benefits: \emph{redundancy}, in case of failure; \emph{improved science acquisition rate} due to parallel operation; and \emph{increased scientific depth} through specialised sensors.

Heterogeneous teams have also been used in analog missions such as the DLR ARCHES experiment on Mount Etna \cite{schuster2020arches} and lava tube exploration for Martian cave analogs \cite{morrell2024robotic}.
The plan for the \acs{NASA} CADRE mission \cite{nayak2025multi} aimed to flight-test a homogeneous swarm for cooperative lunar mapping.

Despite this progress, meaningful cross‑comparison of multi-robot systems remains difficult. 
Most field experiments rely on a small set of scenario‑specific metrics, often inherited from competition scoring rubrics.
Existing \ac{KPI} frameworks also focus on narrow subdomains, such as mapping or \ac{HRI}, and typically do not relate performance to scientific goals.
Hence, there is a need for an easy-to-apply, comprehensive, and science-aligned \ac{KPI} framework that enables cross-project comparison and thereby reveals weaknesses and opportunities for improvement.

In this study, we reverse the conventional approach of deriving \acp{KPI} from engineering field trials.
Instead, we formulate metrics based on scientifically grounded operational scenarios co-developed by lunar geologists and roboticists, thereby ensuring both scientific relevance and operational measurability.
We first outline three conceptual lunar missions targeting the prospecting of (i) ilmenite in volcanic plains, (ii) rare-earth-element-rich ejecta blankets, and (iii) polar water ice. 
We then extract a corresponding set of \acp{KPI} from these scenarios for evaluating heterogeneous multi‑robot exploration strategies.

%% file: chapters/02_related_work.tex
\subsection{Lunar Mission Scenarios}

\noindent Research on surface planetary exploration has proceeded along two largely independent tracks.

\textbf{Robotics-centric efforts} have consistently demonstrated that robots and robotic teams can autonomously explore lunar and Martian analog environments.
Volcanoes \cite{schuster2020arches, burkhard2024collaborative}, deserts \cite{sonsalla2017field}, and lava tubes \cite{brinkmann2024development,morrell2024robotic} are examples of such analog environments.
In many of these studies, the primary focus lies on locomotion and mapping, whereas scientific objectives play a secondary role.
Typical results are 3-dimensional maps \cite{brinkmann2024development, schuster2020arches, morrell2024robotic} or collected samples \cite{schuster2020arches, sonsalla2017field}.
The closest analog mission to a real planetary sampling scenario is the ARCHES Mount Etna campaign \cite{schuster2020arches}, in which pre-selected regions of interest from orbital imagery were first mapped and spectrally classified by an autonomous rover–drone team, then revisited by a second rover that executed geochemical measurements and physical sample acquisition at the identified targets.

\textbf{Science‑driven investigations}, such as \acs{ESA}’s Space Resources Strategy \cite{ESA_SpaceResourcesStrategy_2019}, have identified key lunar resources,  including \ac{REE}, water ice, sunlight, Helium‑3, and ilmenite.
Drawing mainly from \ac{LRO} \cite{nasa_lro} data, researchers have produced increasingly detailed, orbital‑scale datasets that map these resources: radioactive materials, proxies for \ac{REE} \cite{anand2012brief}; evidence for polar water ice \cite{hayne2015evidence}; estimates of regolith Helium‑3 content \cite{fa2007quantitative}; and global TiO$_2$ and FeO abundance maps \cite{sato2017lunar}, proxies for ilmenite.
Landing site ranking frameworks translate this information into kilometer‑scale regions of interest \cite{liu2021landing, kim2022investigation}.
However, these frameworks rarely inform surface mission architectures; traverses, sampling strategies, and task allocations are rarely specified.
Mission-level planning often occurs only after a flight opportunity has been formally identified.
NASA's CADRE project is an example that combines autonomous multi-robot operations with a well-defined science case \cite{nayak2025multi}.

\subsection{Performance Analysis of Multi-Robot Missions}

\noindent Multi-robot field trials are typically evaluated with competition scoring rubrics or subsystem KPIs, including 
\textit{total mapped area} \cite{arm2023scientific, agha2021nebula, sonsalla2017field, schnell2023efficient},
\textit{total distance traveled} \cite{sakagami2023rosmc, agha2021nebula, hudson2022heterogeneous, schnell2023efficient},
\textit{map quality} \cite{agha2021nebula, hudson2022heterogeneous},
\textit{exploration efficiency} \cite{nayak2025multi} and
\textit{total planner‑failure rates} \cite{nayak2025multi}.
The scientific value is measured based on the
\textit{rate of identified resources} \cite{sakagami2023rosmc, arm2023scientific, agha2021nebula, hudson2022heterogeneous, schnell2023efficient} or the \textit{rate of high-resolution images} \cite{morrell2024robotic}.
Most field reports apply only a handful of scenario‑specific engineering KPIs, making cross‑team comparison difficult and leaving science‑driven measures of knowledge gain largely unaddressed.
Comparison frameworks for multi‑robot exploration likewise restrict themselves to 
\textit{map quality} \cite{yan2015metrics, balaguer2009evaluating, arm2023comparison},
\textit{exploration time} \cite{yan2015metrics, xu2022explore} and
\textit{exploration efficiency} \cite{arm2023comparison, xu2022explore},
without links to science objectives.

In the \ac{HRI} community, robotic field trials are evaluated with an emphasis on human factors. 
A survey of 29 studies \cite{murphy2013survey} shows diverse \ac{HRI} metrics, including \textit{time in unscheduled manual operations}, \textit{task success}, and \textit{number of interventions}.
Others are impractical in the field (e.g., estimating an operator's \textit{degree of mental computation}  or \textit{human reliability}), qualitative (e.g., \textit{robot self‑awareness}), or setup‑specific (e.g., \textit{interaction effort} measured as camera‑motion counts).
Human workload is typically assessed by questionnaires \cite{hart1988development}, though recent work explores live psycho-physiological signals \cite{nelles2019evaluation}.

%% file: chapters/03_mission_scenarios.tex
\noindent Given that the primary motivation for lunar prospecting is resource identification, this section introduces three mission scenarios centered on key lunar resources. 
Subsequently, robotic \acp{KPI} are designed to maximize the likelihood of success in these scenarios.
We will first introduce a general mission design that is applicable to various prospecting missions. Subsequently, we propose concrete mission scenarios based on the \acs{ESA} Space Resource Strategy \cite{ESA_SpaceResourcesStrategy_2019}.

\begin{figure}[htbp]
  \centering
  \includegraphics[width=\linewidth]{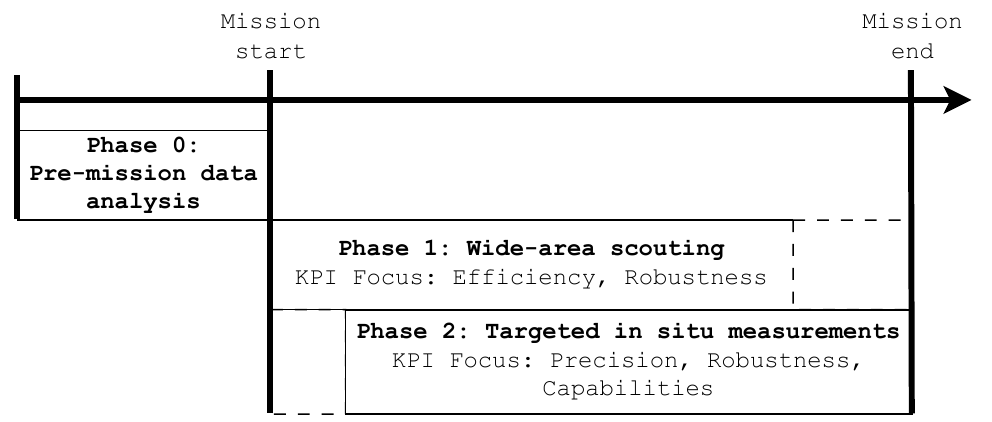}
  \caption{General timeline of the mission concept, indicating relevant KPI focus for each phase.}
  \label{fig:mission_timeline}
\end{figure}

\subsection{Lunar Mission Design}

\noindent We base all scenarios on a three-phase exploration strategy inspired by prior work, including the ARCHES campaign \cite{schuster2020arches} and Team GLIMPSE \cite{arm2023scientific} from the Space Resources Challenge \cite{esa_src_challenge2021}, with a visual overview shown in Figure~\ref{fig:mission_timeline}.
Before launch, \textbf{Phase 0} consists of analyzing orbital satellite data, e.g., \ac{LRO} \cite{nasa_lro} via the Lunar QuickMap \cite{lroc_quickmap}, to obtain a coarse estimate of local resource abundances and define a preliminary prospecting grid.
In \textbf{Phase 1}, scouting robots (\emph{scouts}) map the prospecting area remotely using optical and spectral cameras tailored for the resource of interest, refining the initial prospecting grid based on updated surface observations.
\textbf{Phase 2} follows with high-resolution, targeted in situ measurements by specialized robots (\emph{scientists}), who deploy scientific instruments at strategic locations for precise elemental analysis, including subsurface measurements using robotic arms or drills.

\subsection{Scenario 1: Ilmenite Exploration}

\noindent Ilmenite (\textit{FeTiO$_3$}) is a promising, quasi-globally abundant source of oxygen for life support and an oxidizer for propulsion \cite{crawford2015lunar}.
While the horizontal extent of ilmenite can be inferred from its proxies TiO$_2$ and FeO, the vertical extent is largely unknown \cite{giguere2020volcanic}.
The mission goal is therefore to (1) characterize vertical profiles using impact craters as natural windows into deeper layers, and (2) validate horizontal distributions with in situ measurements for cross-correlation with orbital maps.

% We analyzed candidate lunar sites by starting from ilmenite‑rich regions proposed by Gaddis et al. \cite{gaddis2003compositional} and reviewing TiO$_2$/FeO maps in the Lunar QuickMap. 
We selected the Sinus Aestuum region, which shows a clear, continuous gradient in both proxies (Figure~\ref{fig:scenario1}e,f), with a landing zone at low slope and rock abundance enabling a mission lasting one lunar day ($\sim$14 Earth days).
Measurements must be taken along the horizontal and vertical elemental gradient to meet scientific objectives.
The area contains impact craters up to $\sim$\SI{1}{\kilo\meter} in diameter, enabling subsurface sampling at multiple depths, though accessing them requires traversing rocky slopes of up to \SI{25}{\degree}.
% From the landing zone, scouts with spectral cameras rapidly assess the terrain to refine the exploration strategy. Scientists then acquire elemental compositions along the prospecting grid using \acp{APXS}, \acp{LIBS}, or \acp{XRF} at a \SI{50}{\meter} sampling interval.
This consideration leads to the following exploration strategy, which is also evident in the marked preliminary prospecting trajectory in Figure~\ref{fig:scenario1}a.
From the landing zone, scouts equipped with, e.g., spectral cameras rapidly assess the terrain to refine the exploration strategy for the scientists.
Subsequently, scientists acquire detailed elemental compositions along the marked lines using \acp{APXS}, \acp{LIBS}, or \acp{XRF}. Since these measurements require the robot to be stationary, we propose a \SI{50}{\meter} sampling interval to match the resolution of the orbital FeO map. 

\subsection{Scenario 2: \acf{REE} Exploration}

\begin{figure}
    \centering
    \includegraphics[width=.9\linewidth]{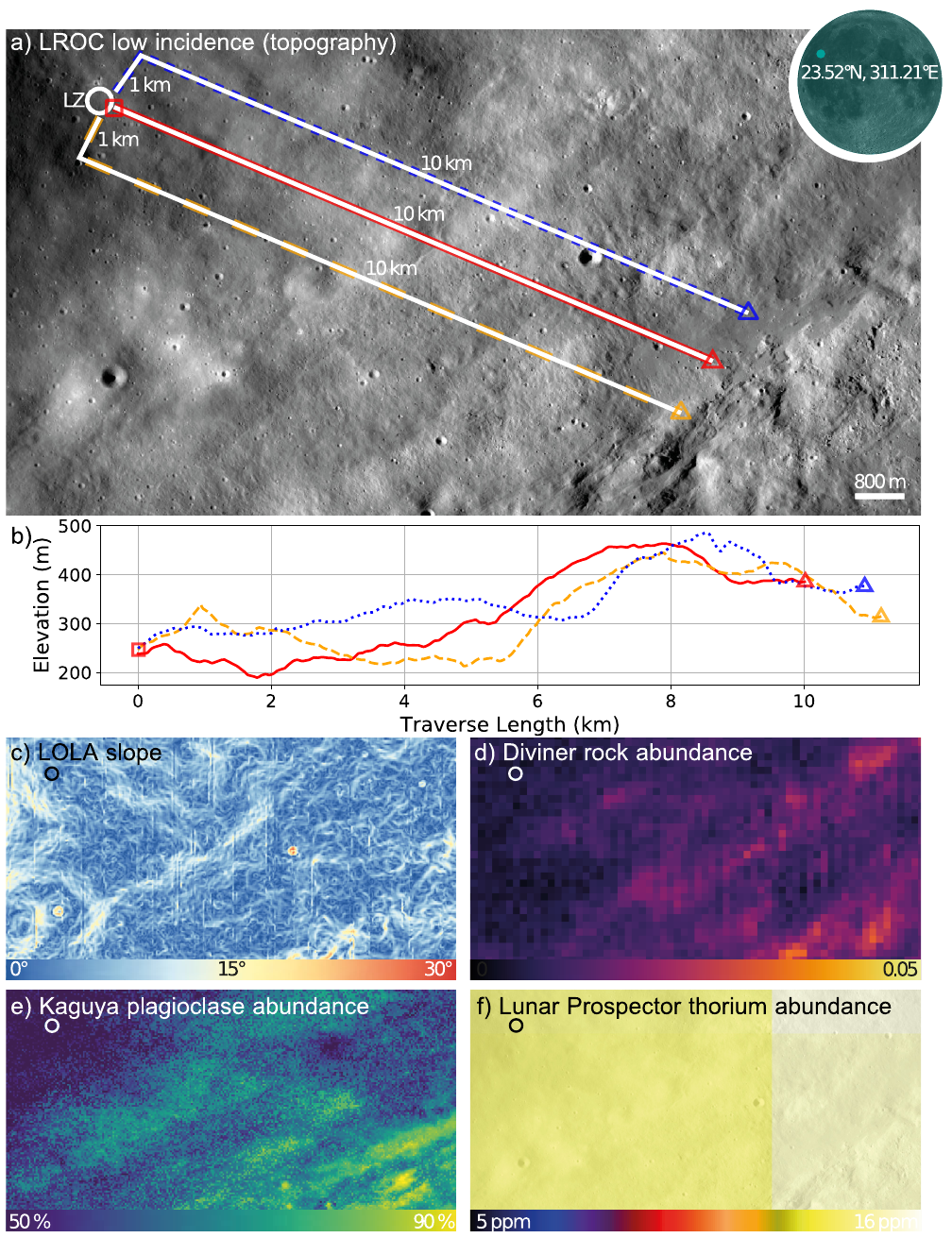}
    \caption{Scenario 2 (KREEP). (a) LROC low-incidence image mosaic with landing zone (LZ), preliminary prospecting grid (white), and elevation transect (blue, red, orange). (b) Elevation profile along the transect. (c–f) Selected orbital maps showcasing thermophysical, topographic, and compositional properties, with the LZ indicated by a white circle.}
    \label{fig:scenario2}
\end{figure}

\begin{figure}
    \centering
    \includegraphics[width=.9\linewidth]{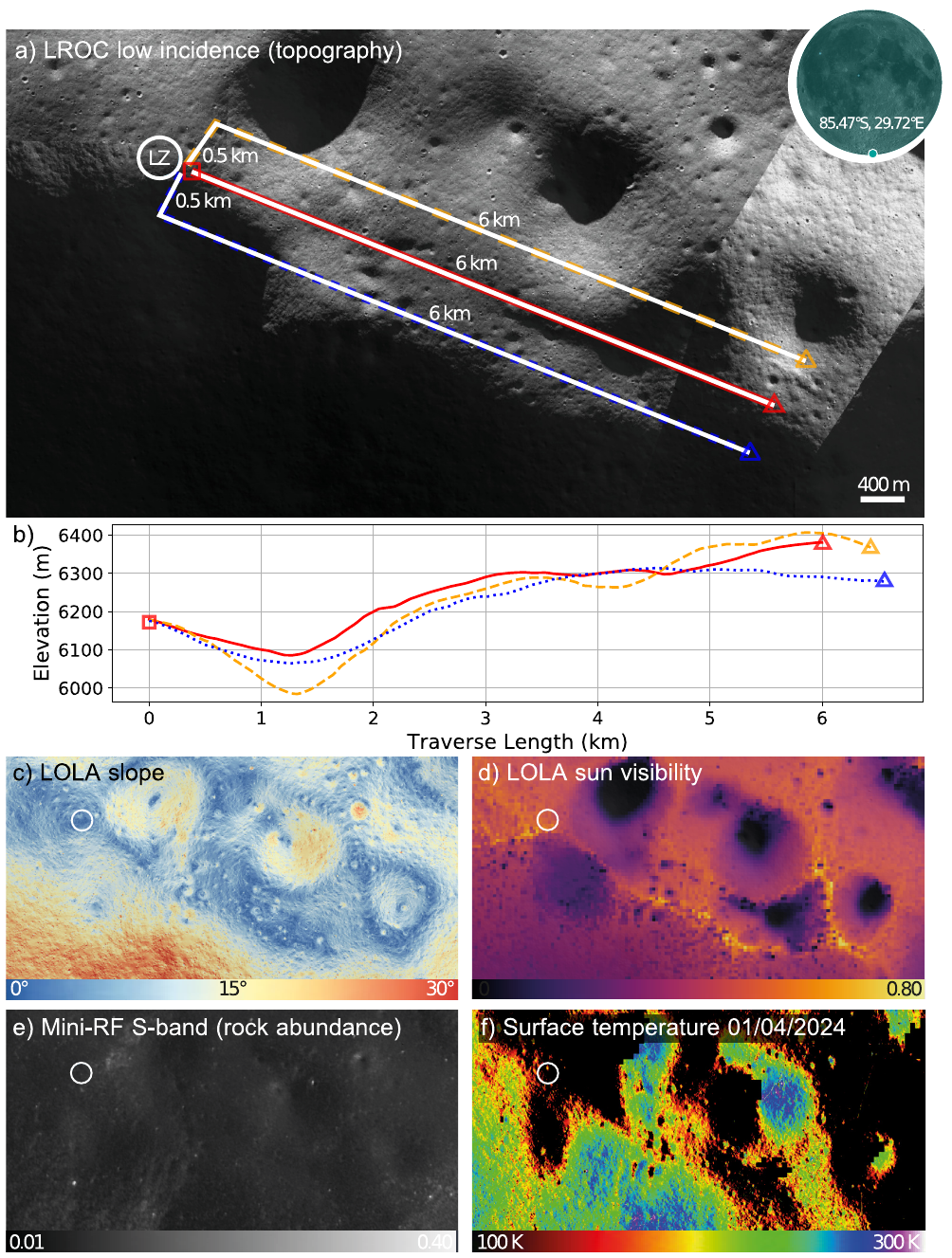}
    \caption{Scenario 3 (Water ice). (a) LROC low-incidence image mosaic with landing zone (LZ), preliminary prospecting grid (white), and elevation transect (orange, red, blue). (b) Elevation profile along the transect. (c–f) Selected orbital maps showcasing thermophysical and topographic properties, with the LZ indicated by a white circle.}
    \label{fig:scenario3}
\end{figure}

\noindent \acp{REE} are valuable for technologies such as electronics, magnets, and batteries, and their in situ supply on the Moon could facilitate lunar infrastructure. They are concentrated in KREEP material, which contains radioactive, heat-producing elements (e.g., thorium) detectable via remote sensing \cite{anand2012brief}. Their vertical extent and internal nature, however, remain poorly characterized, constraining \ac{ISRU} assessments \cite{ligeza2025exploring}.

% Based on the orbital proxies, we identified several KREEP hotspots on the lunar nearside using Lunar QuickMap.
Our scenario targets a prominent KREEP site on the Aristarchus Plateau (Figure~\ref{fig:scenario2}). 
The terrain consists primarily of rocky ejecta with slope angles up to \SI{25}{\degree}, and line-of-sight communication may be hindered by boulders and topographic undulations.
We propose an initial exploration strategy along the white lines in Figure~\ref{fig:scenario2}a, sampling along the ejecta of the Aristarchus crater, as indicated by the rock abundance in Figure~\ref{fig:scenario2}d. 
Scouts equipped with gamma-ray and neutron spectrometers remotely detect thorium 
and local KREEP hotspots \cite{ligeza2025exploring}, which scientists then 
investigate in situ using \ac{XRF} and laser-induced mass spectrometry.

\subsection{Scenario 3: Water Ice Characterization in Polar \acsp{PSR}}

\input{chapters/04_kpi_definition_table}

\noindent Scenario 3 focuses on characterizing water ice — a key resource for life support and propellant production \cite{crawford2015lunar}. 
Water ice is stable only in \acp{PSR} near the lunar poles, but its precise distribution, physical form, and abundance remain unknown \cite{crawford2015lunar}.
Prospecting missions must therefore not only detect water ice but also characterize its thermophysical properties to assess extraction feasibility.

We target the mountainous terrain west of the Nobile Crater, the same region as NASA's VIPER mission \cite{colaprete2021volatiles}, chosen for its varied PSR morphologies and predicted thermally stable zones (Figure~\ref{fig:scenario3}f). Accessing specific \acp{PSR} may require traversing slopes up to \SI{20}{\degree}, and crater descents may require advanced communication planning.
The exploration strategy with the marked preliminary prospecting trajectory for this scenario is shown in Figure~\ref{fig:scenario3}a.
Scouts carry spectral cameras, neutron spectrometers, and \acp{GPR} to identify water-bearing materials and subsurface deposits. Scientists investigate targets in situ with \ac{LIBS} and mass spectrometry for isotopic compositions.

%% file: chapters/04_kpi_definition_table.tex
%––––––––––  KPI TABLE  ––––––––––
\begin{table*}[!t]
  \caption{Key Performance Indicators (KPIs) for Multi-Robot Lunar Missions. $\downarrow$ = lower value is better, $\uparrow$ = higher value is better.}
  \label{tab:kpis}
  \centering
  \footnotesize
  \setlength{\tabcolsep}{3pt}
  \renewcommand{\arraystretch}{1.1}
  \begin{tabularx}{\textwidth}{@{}>{\RaggedRight\arraybackslash}p{1.7cm} K{4cm} l X@{}}
    \toprule
    \textbf{Category} & \textbf{KPI} & \textbf{Unit} & \textbf{Definition} \\ \midrule[1pt]

    %––––– Efficiency –––––
    \multirow{7}{=}{Efficiency}
        % & Total Explored Area $\uparrow$ & \si{\meter\squared} & Cumulative area mapped: $A_{\text{mapped}}$ \\ \cmidrule(lr){2-4}
        % & Total Distance Traveled $\uparrow$ & \si{\meter} & Total distance traveled by all robots: ${d_{\text{tot}}}$ \\ \cmidrule(lr){2-4}
        & Mapping Efficiency $\uparrow$ & \si{\meter\squared\per\meter} & Area mapped per meter traveled (total explored area / total distance traveled by all robots): ${A_{\text{mapped}}}/{d_{\text{tot}}}$ \\ \cmidrule(lr){2-4}
        & Mapping Rate $\uparrow$ & \si{\meter\squared\per\second} & Area mapped per unit time (total explored area / mission time): ${A_{\text{mapped}}}/{t_{\text{mission}}}$ \\ \cmidrule(lr){2-4}
        & Task Success Ratio $\uparrow$ & \% & Percentage of assigned tasks completed: ${N_{\text{completed}}}/{N_{\text{total}}}\times100$ \\ \cmidrule(lr){2-4}
        & Subjective Operator Workload $\downarrow$ & -- & Subjective operator workload assessed via standardized scales (e.g., NASA‑TLX) or physiological proxies (e.g., pulse rate) \\ \cmidrule(lr){2-4}
        & Quantitative Operator Workload $\downarrow$ & \% & Percentage of mission time spent interacting with the robots: ${t_{\text{operator}}}/{t_{\text{mission}}}\times100$\\ \midrule[1pt]

    %––––– Robustness –––––
    \multirow{5}{=}{Robustness}
        & Robot Downtime $\downarrow$ & \% & Percentage of mission time the robot cannot execute planned actions (e.g., waiting for commands or recovering from faults): ${t_{\text{idle}}}/{t_{\text{mission}}}\times100$ \\ \cmidrule(lr){2-4}
        & Autonomy Ratio $\uparrow$ & -- & Fraction of mission the robot operates without human intervention: $1-\text{RAD}$, with $\text{RAD}=t_\text{IE}/(t_\text{IE}+t_\text{NT})$, 
        where $\text{IE}$ is the interaction effort and $\text{NT}$ is the neglect tolerance
        \\ \cmidrule(lr){2-4}
        & Time in Unscheduled Manual Operations $\downarrow$ & \% & Percentage of mission time spent in unexpected teleoperation or manual override modes: ${t_{\text{unscheduled-manual}}}/{t_{\text{mission}}}\times100$ \\ \cmidrule(lr){2-4}
        & Retry Ratio $\downarrow$ & \% & Percentage of task attempts that were retries: $({N_{\text{attempts}}-N_{\text{success}}})/{N_{\text{attempts}}}\times100$ \\ \midrule[1pt]
        
    %––––– Precision –––––
    \multirow{5}{=}{Precision}
        & Science Acquisition Density $\uparrow$ & \si{\per\meter\squared} & Number of valid scientific measurements per mapped area: $N_{\text{measurements}}/A_{\text{mapped}}$ \\ \cmidrule(lr){2-4}
        & Science Acquisition Distribution $\uparrow$ & -- & Clark–Evans nearest‑neighbor ratio $R$ of measurement locations ($R{=}1$ random, $R{<}1$ clustered, $R{>}1$ dispersed) \\ \cmidrule(lr){2-4}
        & Localization Error $\downarrow$ & \si{\meter} & Pose estimation error (e.g., RMSE of Absolute Trajectory Error for position) \\ \cmidrule(lr){2-4}
        & Instrument Placement Error $\downarrow$ & \si{\meter} & Euclidean distance between commanded and achieved instrument locations (median or RMSE) \\ \cmidrule(lr){2-4}
        & Remote Sensing Error $\downarrow$ & \si{\meter} & Euclidean distance between remotely sensed target and ground-truth measurements (e.g., RMSE over targets) \\ \cmidrule(lr){2-4}
        & Map Error $\downarrow$ & \si{\meter} & Map-ground‑truth discrepancy (e.g., bidirectional Chamfer distance): $d_{\text{Chamfer}}(M,\;GT)$ \\ \cmidrule(lr){2-4}
        & Ratio of Identified Resources $\uparrow$ & \% & Percentage of ground‑truth resources correctly detected: ${N_{\text{resource}}}/{N_{\text{GT}}}\times100$ \\ %\midrule[1pt]

    % %––––– Capabilities –––––
    % \multirow{4}{=}{Capabilities}
    %     & Maximum Traversable Slope $\uparrow$ & \si{\deg} & Steepest slope the robot traverses without failure (ascent and descent reported separately) \\ \cmidrule(lr){2-4}
    %     & Terrain Adaptability for Deployment $\uparrow$ & \si{\deg} & Largest ground slope on which the robot can deploy instruments within instrument stability tolerances (no slip, tilt  error$\leq\theta_{\text{tol}}$, placement error $\leq e_{\text{tol}}$) \\ \cmidrule(lr){2-4}
    %     & Maximum Speed $\uparrow$ & \si{\meter\per\second} & Highest sustained travel speed over $\geq5~m$ on flat, even terrain without failure \\ \cmidrule(lr){2-4}
    %     & Energy Consumption $\downarrow$ & \si{\joule\per\meter} & Mean energy per meter traveled (mission average on different terrains) \\
    \bottomrule
  \end{tabularx}
  \vspace{-1.3em}
\end{table*}
%––––––––––––––––––––––––––––––––––––––––––––––––––

%% file: chapters/04_kpi_definition.tex
\noindent Lunar resource prospecting missions place three demands:
\begin{enumerate}
    \item Cover a large area in a limited time,
    \item Continue operation productively under unexpected or changing conditions,
    \item Obtain and maximize scientific results.
    % \item Physically reach and operate in scientifically relevant locations.
\end{enumerate}
We propose a \ac{KPI} framework mapping each demand to a category: \textbf{Efficiency} (Req.~1), \textbf{Robustness} (Req.~2), and \textbf{Precision} (Req.~3), with formal definitions in Table~\ref{tab:kpis}.
% To make these requirements measurable, we propose a \ac{KPI} framework that maps each demand to a corresponding category: \textbf{Efficiency} (Req.~1), \textbf{Robustness} (Req.~2), and \textbf{Precision} (Req.~3). %, and \textbf{Capabilities} (Req.~4).
% All formal definitions are provided in Table~\ref{tab:kpis}.

\textbf{Efficiency} evaluates how well agents convert time, distance traveled, and operator attention to the mapped area and completed tasks, especially during scouting in \textit{Phase 1}.
For \textit{S1 (Ilmenite)} and \textit{S2 (KREEP)}, broad areal coverage is essential to sample varying elemental concentrations along the gradient within the $\sim$14~day illumination window.
While the total explored area is a primary mission outcome, it is typically dictated by the mission definition and therefore cannot be used to compare efficiency across different missions.
Instead, we propose measuring \textit{Mapping Efficiency} (area per meter travelled), and \textit{Mapping Rate} (area per second).
In order to calculate these values, the total area $A_{mapped}$, the total distance travelled $d_{tot}$, and the total mission duration $t_{mission}$ are needed.
All three quantities can be easily obtained by analysing the final created map and the recorded robot trajectories.

For \textit{S3 (Water ice)}, the goal shifts from full coverage to successfully characterizing chosen \acp{PSR}.
This requires completing high-risk tasks, such as crater descents and climb-outs, which are captured by the \textit{Task Success Ratio}.
To measure this value, task types must be clearly defined, and a consistent method for task logging must be implemented.
While post-mission analysis of robot behavior is one option, we recommend explicit state logging that records when a task is assigned, when execution begins, and whether the robot successfully completes it.

Across all scenarios, human time is a scarce resource.
High operator workload reduces scalability, increases error rates, and slows decision-making. 
Standardized questionnaires (e.g., NASA‑TLX) and physiological proxies (e.g., pulse rate) can indicate \textit{Subjective Operator Workload}, but they are either retrospective or strongly biased by the overall stress level typical for mission operations.
We therefore complement these with a mission‑integrated \textit{Quantitative Operator Workload}, defined as the percentage of mission time humans spend operating the robots, which more directly reflects actual staffing needs during operations.
To calculate this value, the total time the operator interacted with the robots, $t_{operator}$, must be measured along with the total mission time, $t_{mission}$.
% This can be achieved either by implementing a logging system that detects operator-issued actions (e.g., teleoperation commands or button presses for task allocation) or by recording the operator and their screen and extracting interaction intervals through post-mission analysis.

\textbf{Robustness} evaluates how reliably the system maintains productive operation when conditions deviate from the plan, be it due to communication dropouts, unexpected terrain, or task failures.
In \textit{S2 (KREEP)}, long traverses ($\sim$\SI{11}{\kilo\meter}) make \textit{Robot Downtime} a key limitation, as any idle period equates to lost coverage.
It is measured as the percentage of mission time during which a robot did not contribute productively to mission progress, i.e., the ratio between its idle time $t_{idle}$ (e.g., waiting for tasks or planning) and the total mission duration $t_{mission}$.

In \textit{S3 (Water ice)}, the potential for fragile communication inside \acp{PSR} necessitates high autonomy. 
We quantify this with the \textit{Autonomy Ratio}—defined as the inverse of Robot Attention Demand (RAD) \cite{arm2023comparison}—and with \textit{Time in Unscheduled Manual Operations}. 
The RAD is defined as
\begin{equation}
    RAD = \frac{t_{IE}}{t_{IE}+t_{NT}}
\end{equation}
where $t_{IE}$ is the interaction effort, i.e., the time spent per operator interaction, and $t_{NT}$ is the neglect tolerance, i.e., the time the robot remains productive without operator input. 
The \textit{Time in Unscheduled Manual Operations} is defined as the percentage of mission time during which robots require unexpected manual override, measured as $t_{\text{unscheduled-manual}}/t_{\text{mission}}$.
For evaluation of both metrics, explicit state logging—as introduced above—allows extraction of manual commands issued to the robots (e.g., teleoperation inputs) and the corresponding autonomous activities such as planning and execution.
Given the strict mission time budgets and the elevated risk of certain tasks, such as crater descents, robustness is further captured by the \textit{Retry Ratio}, which quantifies the fraction of task attempts that were retries.

\textbf{Precision} describes the scientific value of a mission by measuring the accuracy of locating, identifying, and quantifying resources.
For \textit{S1 (Ilmenite)} and \textit{S2 (KREEP)}, which aim to map horizontal distributions, we emphasize taking ``enough" and ``well‑spread" samples: \textit{Science Acquisition Density} (samples per area) and \textit{Science Acquisition Distribution}.
For the latter, we propose using the Clark–Evans nearest‑neighbor ratio $R$ \cite{clark1954distance}, which compares the observed mean nearest‑neighbor distance to the value expected under a Poisson distribution representing complete spatial randomness, targeting $R{>}1$ (over-dispersion, i.e., a more uniform spacing than random). 
Both metrics can be computed directly from logged sample coordinates.

Accuracy in where and what we measure is equally critical.
We therefore include \textit{Localization Error} (robot pose), \textit{Instrument Placement Error} (where the tool touches down), and \textit{Remote Sensing Error} (locating scientific targets by non-contact sensors, e.g., spectral cameras).
Finally, accurate 3D terrain reconstructions from the scouting in \textit{Phase 1} are essential for the science in \textit{Phase 2}.
Since path planning and coordination rely on map fidelity, we additionally include \textit{Map Error}.

For \textit{S2 (KREEP)} and \textit{S3 (Water ice)}, where targeted resources are sparse, mission success depends on reliable detection and characterization. We therefore measure the \textit{Ratio of Identified Resources}, which can be obtained in analog scenarios by comparing the set of detected resources against the known ground-truth resource locations.

% \textbf{Capabilities} express platform limits that bound where and how science can be done, even if they do not directly measure mission success.
% All three scenarios require access to the interior of the crater.
% Thus, \textit{Maximum Traversable Slope} and \textit{Terrain Adaptability for Deployment} (using instruments on inclined ground) are essential.
% Furthermore, long traverses in \textit{S2 (KREEP)} and \textit{S3 (Water ice)} benefit from higher \textit{Maximum Speed} and lower \textit{Energy Consumption} per meter, which directly extend spatial reach for a fixed energy budget.

% These capability metrics can be obtained through a combination of onboard telemetry and post-mission terrain analysis. In practice, \textit{Maximum Traversable Slope} and \textit{Terrain Adaptability for Deployment} can better be characterized separately from the field test in dedicated trials, while \textit{Maximum Speed} and \textit{Energy Consumption} per meter can be computed directly from logged velocity and power measurements during the mission.

%% file: chapters/05_discussion.tex
\subsection{Relevance of Proposed \acsp{KPI}}
\noindent The proposed KPI framework is designed to systematically evaluate heterogeneous robotic teams in lunar prospecting scenarios. However, the relative importance of individual \acp{KPI} varies with the objectives and constraints of each scenario; experiments may select the subset that best aligns with their mission goals, as summarized in Table~\ref{tab:kpi_scenario_phase_check}.

\subsubsection{Scalability of KPIs to Multi-Robot Teams}
While the proposed \acp{KPI} are motivated by multi-robot missions, many of them are not tied to multi-robot coordination itself, but rather quantify general aspects of mission performance.
Nevertheless, all \acp{KPI} are defined at the team level and can be directly applied to heterogeneous multi-robot systems by aggregating robot-level measurements and relating them to shared mission objectives.
In this way, the framework remains applicable to both single- and multi-robot deployments, while explicitly capturing the benefits and challenges that emerge when multiple robots operate in parallel.

\subsubsection{Relevance of KPI Set}
\textit{Efficiency} \acp{KPI} capture the effectiveness of resource utilization in exploration. For \textit{S1 (Ilmenite)} and \textit{S2 (KREEP)}, dense and rapid mapping (E.1, E.2) is essential, especially for the scouting phase. In contrast, \textit{S3 (Water ice)} prioritizes detailed characterization within selected PSRs rather than extensive area coverage, making broad area efficiency less critical.

\textit{Robustness} \acp{KPI} quantify how reliably robotic systems operate under challenging conditions. In \textit{S2}, where long traverses are necessary, minimizing \textit{Robot Downtime} (R.1) is crucial. \textit{S2} and \textit{S3}, which involve unstable or limited communication, require higher autonomy, making \textit{Autonomy Ratio} (R.2) and \textit{Time in Unscheduled Manual Operations} (R.3) key indicators of operational robustness.

\textit{Precision} \acp{KPI} directly link to scientific quality, emphasizing accurate identification, localization, and quantification of lunar resources. The \textit{Ratio of Identified Resources} (P.7) is particularly relevant for \textit{S2} and \textit{S3}, where targeted resources are sparse, requiring reliable detection and characterization. Conversely, S1 emphasizes correlating in situ data with orbital measurements, making the measurement distribution (P.1, P.2) and localization metrics (P.3, P.6) more significant.

\begin{table}[htbp]
\caption{Relevance of each KPI for scenario–phase combinations. 
Phase 1 (p1) refers to wide-area scouting, Phase 2 (p2) to targeted in situ operations. 
KPIs are classified in: \protect\fullcirc \space highly relevant; \protect\halfcirc relevant; \protect\emptycirc \space not relevant.}
\label{tab:kpi_scenario_phase_check}
\centering
\footnotesize
\renewcommand{\arraystretch}{1.2}
\resizebox{\linewidth}{!}{%
\begin{tabular}{@{}>{\raggedright\arraybackslash}p{4.5cm} *{6}{>{\centering\arraybackslash}p{0.28cm}}@{}}
\toprule
\textbf{KPI} 
& \multicolumn{2}{c}{\textbf{S1}} 
& \multicolumn{2}{c}{\textbf{S2}} 
& \multicolumn{2}{c}{\textbf{S3}} \\
\cmidrule(lr){2-3} \cmidrule(lr){4-5} \cmidrule(l){6-7}
& \textbf{p1} & \textbf{p2} 
& \textbf{p1} & \textbf{p2} 
& \textbf{p1} & \textbf{p2} \\
\midrule
E.1 Mapping Efficiency & \fullcirc & \halfcirc & \fullcirc & \halfcirc & \emptycirc & \emptycirc \\
E.2 Mapping Rate & \fullcirc & \halfcirc & \fullcirc & \halfcirc & \emptycirc & \emptycirc \\
E.3 Task Success Ratio & \fullcirc & \fullcirc & \fullcirc & \fullcirc & \fullcirc & \fullcirc \\
E.4 Subjective Operator Workload & \fullcirc & \fullcirc & \fullcirc & \fullcirc & \fullcirc & \fullcirc \\
E.5 Quantitative Operator Workload & \fullcirc & \fullcirc & \fullcirc & \fullcirc & \fullcirc & \fullcirc \\
\midrule
R.1 Robot Downtime & \halfcirc & \halfcirc & \fullcirc & \fullcirc & \emptycirc & \emptycirc \\
R.2 Autonomy Ratio & \halfcirc & \halfcirc & \fullcirc & \fullcirc & \fullcirc & \fullcirc \\
R.3 Time in Unscheduled Manual Ops & \halfcirc & \halfcirc & \fullcirc & \fullcirc & \fullcirc & \fullcirc \\
R.4 Retry Ratio & \fullcirc & \fullcirc & \fullcirc & \fullcirc & \fullcirc & \fullcirc \\
\midrule
P.1 Science Acquisition Density & \halfcirc & \fullcirc & \halfcirc & \halfcirc & \halfcirc & \halfcirc \\
P.2 Science Acquisition Distribution & \halfcirc & \fullcirc & \halfcirc & \halfcirc & \halfcirc & \halfcirc \\
P.3 Localization Error & \fullcirc & \fullcirc & \halfcirc & \halfcirc & \halfcirc & \halfcirc \\
P.4 Instrument Placement Error & \emptycirc & \fullcirc & \emptycirc & \fullcirc & \emptycirc & \fullcirc \\
P.5 Remote Sensing Error & \fullcirc & \fullcirc & \fullcirc & \fullcirc & \fullcirc & \fullcirc \\
P.6 Map Error & \fullcirc & \fullcirc & \halfcirc & \halfcirc & \halfcirc & \halfcirc \\
P.7 Ratio of Identified Resources & \halfcirc & \halfcirc & \halfcirc & \fullcirc & \halfcirc & \fullcirc \\
% \midrule
% C.1 Max Traversable Slope & \emptycirc & \halfcirc & \emptycirc & \fullcirc & \emptycirc & \fullcirc \\
% C.2 Terrain Adaptability & \emptycirc & \halfcirc & \emptycirc & \fullcirc & \emptycirc & \fullcirc \\
% C.3 Max Speed & \halfcirc & \halfcirc & \fullcirc & \fullcirc & \fullcirc & \fullcirc \\
% C.4 Energy Consumption & \halfcirc & \halfcirc & \fullcirc & \fullcirc & \fullcirc & \fullcirc \\
\bottomrule
\end{tabular} }
\end{table}

% \subsubsection{Capabilities}
% Capability \acp{KPI} define the robotic platform's limits that affect the operational feasibility. All scenarios require careful consideration of robot capabilities, but they become especially pertinent for \textit{S2}'s uneven terrain and \textit{S3}'s challenging \ac{PSR}s. 

\subsection{Advantages of Scenario-Based \acs{KPI} Development}
\noindent Grounding \ac{KPI} selection in realistic mission scenarios highlights the practical value of each metric compared to previous frameworks. Traditional metrics, such as \textit{Total Distance Traveled}, offer limited insights unless directly linked to scientific outcomes, making \textit{Mapping Efficiency} a more meaningful comparative measure.

Previous studies often prioritized \textit{Task Success Ratio}, but this framework also emphasizes \textit{Retry Ratio} to capture inefficiencies arising from task repetition specifically. Metrics such as \textit{Ratio of Identified Resources}, crucial in resource-sparse contexts, are scenario-dependent and may not universally reflect mission success, exemplified by \textit{S1 (Ilmenite)}.

In contrast, previously omitted metrics, such as \textit{Instrument Placement Error} and \textit{Remote Sensing Error}, reflect the reliability and scientific relevance of the gathered data, thereby providing valuable insights.
Furthermore, traditional terrestrial metrics such as \textit{Robot Downtime} need to be redefined in lunar contexts, where strategic downtime (e.g., thermal management, solar charging) may be deliberate and beneficial rather than inefficient.

\subsection{Feasibility and Practicality of \acs{KPI} Measurement}
% \begin{figure}
%     \centering
%     \includegraphics[width=\linewidth]{fig/field_test_team.jpg}
%     \caption{Heterogeneous robotic team during the lunar-analog field deployment used to assess the practical applicability of the proposed \ac{KPI} framework.}
%     \label{fig:field_test}
% \end{figure}

\noindent Ensuring that \acp{KPI} can be practically measured in terrestrial field tests is crucial for their applicability and was thus a main objective of this work. 
We therefore evaluated the feasibility of KPI extraction in a multi-robot field deployment, which implemented a lunar-analog prospecting scenario focused on detecting and mapping discrete boulders and soil patches. 
The team consisted of one operator and five robots: three scouts and two scientists.
In contrast to the here introduced scenarios, where broad areal mapping and spatial coverage are primary objectives, our field trial emphasized reliably finding and confirming sparse resources, making it most comparable to \textit{S2 (KREEP)} in terms of search and verification behavior.
%%% TODO: needs to be anonymised in case submitted
Details of the deployment, methodology, and the resulting \ac{KPI} values are reported in the corresponding paper~\cite{oberacker2026MOSAIC}.
Here, we focus on assessing the practicality of applying the proposed \ac{KPI} framework in a real-world setting.

% Robustness:
Overall, the time- and activity-based \textbf{Robustness} metrics (\textit{Robot Downtime}, \textit{Autonomy Ratio}, \textit{Time in Unscheduled Manual Operations}, and \textit{Retry Ratio}) can be extracted reliably when the mission software provides explicit state logging for task assignment, execution, and completion, as well as for manual interventions.
In our deployment, we used a behavior-tree–based control system that tracked task assignment and execution, enabling the reconstruction of a detailed timeline of each robot’s activities. 
As a result, all timestamps required to compute these \acp{KPI} were available.

% Efficiency
Operator workload metrics require careful integration into the mission setup. While \textit{Subjective Operator Workload} is inherently difficult to assess objectively in field conditions—since field tests are generally stressful and exhausting, particularly when the operator is part of the development team—we decided against using this measure and instead focused on \textit{Quantitative Operator Workload}.
Although this metric is in principle measurable, it is non-trivial to implement robustly. 
In our field test, we estimated it through post-mission video annotation of the operator and screen recordings, which proved feasible but noisy. 
We therefore recommend implementing dedicated logging of operator-issued actions (e.g., teleoperation commands, task dispatches, and autonomy mode overrides) to directly compute interaction time. 
In particular, automatically capturing interactions performed through RViz can be challenging, as they may correspond to diverse UI actions that do not consistently map to a single logged robot command.
The remaining \textbf{Efficiency} \acp{KPI}, namely \textit{Mapping Efficiency}, \textit{Mapping Rate}, and \textit{Task Success Ratio}, were comparatively easy to measure based on the produced 3D map, the recorded traversal distances of the robots, and manually annotated task success.

% Precision
\textit{Science Acquisition Density} and \textit{Science Acquisition Distribution} were not evaluated in our field trial, as the mission objective focused on resource identification rather than dense sampling for spatial distribution mapping, rendering these \textbf{Precision} \acp{KPI} irrelevant in this context.
Moreover, the metrics \textit{Localization Error}, \textit{Instrument Placement Error}, and \textit{Remote Sensing Error} rely on accurate ground-truth data (e.g., surveyed target locations or reference maps), which were not available for our deployment.
In contrast, the \textit{Ratio of Identified Resources} can be readily evaluated as long as the number and approximate locations of the ground-truth resources are known.

In summary, the proposed \ac{KPI} framework proved easy to apply for \textbf{Efficiency} and \textbf{Robustness} metrics, provided that the mission software includes sufficiently detailed state and interaction logging. 
Most time- and activity-based \acp{KPI} can be extracted directly from such logs with minimal additional instrumentation. 
In contrast, \textbf{Precision} metrics require substantially more preparation, as they depend on the availability of reliable ground-truth data, such as surveyed maps or reference target locations, which is not always feasible in outdoor analog field tests. 
This highlights the importance of aligning the selected \acp{KPI} with both the mission objectives and the practical constraints of the experimental setup.

\subsection{Limitations and \acs{KPI} Trade-offs}
\noindent Several proposed \acp{KPI} are correlated, which may require thoughtful interpretation during analysis. For instance, \textit{Mapping Efficiency}, \textit{Total Explored Area}, and \textit{Total Distance Traveled} all relate to coverage; however, \textit{Mapping Efficiency} provides the most meaningful basis for cross-comparison between different missions. Similarly, while \textit{Subjective} and \textit{Quantitative Operator Workload} assess operator burden, subjective assessments may be more insightful but more complex to measure consistently.
Metrics such as \textit{Retry Ratio} and \textit{Task Success Ratio} reflect task reliability. Still, the former emphasizes inefficiency caused by task repetition, while the latter focuses on the ultimate completion of the task.
Trade-offs between \acp{KPI} must also be explicitly considered in mission design. 
Optimizing \textit{Science Acquisition Density} typically lowers \textit{Explored Area}, and high \textit{Mapping Rates} may compromise precision (e.g., \textit{Map Error} or \textit{Ratio of Identified Resources}). 
Thus, selecting optimal \ac{KPI} targets must be done with respect to the specific context of the scenario, taking into account scientific priorities, terrain conditions, and available resources. This ensures system performance is evaluated meaningfully while acknowledging the trade-offs between efficiency, robustness, and data quality in real mission scenarios.

%% file: chapters/06_conclusion.tex
\noindent This paper addresses the challenge of robotic resource prospecting in extraterrestrial environments by introducing three representative lunar mission scenarios targeting ilmenite, rare-earth elements, and water ice. 
Based on these scenarios, we propose a structured \ac{KPI} framework that systematically evaluates the performance of heterogeneous robotic teams while explicitly accounting for scenario-specific objectives and operational constraints. 
As a result, the relative importance of individual \acp{KPI} naturally varies across scenarios and mission phases, reflecting different scientific goals and risk profiles.

We validated the proposed framework in a multi-robot field deployment and found it straightforward to apply to Efficiency and Robustness metrics, whereas Precision metrics require ground-truth data, which is not always feasible in outdoor settings. 
%We validated the proposed framework in a multi-robot field deployment and found it straightforward to apply to efficiency- and robustness-related \acp{KPI}, provided that appropriate mission-state and interaction logging is available. 
% In contrast, precision-oriented \acp{KPI} require preparation, such as reliable ground-truth data, which is not always feasible in outdoor analog environments. 
While most \acp{KPI} quantify general mission performance rather than explicit inter-robot coordination, they are defined at the team level and naturally extend to heterogeneous multi-robot systems through aggregation and shared mission objectives.
Overall, the framework provides a clear and practical basis for evaluating heterogeneous robotic teams, enabling meaningful cross-mission comparisons, informed mission design, and targeted technological development for future lunar exploration.